\pdfoutput=1

\documentclass[11pt]{article}

\usepackage[preprint]{acl}

\usepackage{times}
\usepackage{latexsym}

\usepackage[T1]{fontenc}

\usepackage[utf8]{inputenc}

\usepackage{microtype}

\usepackage{inconsolata}

\usepackage{graphicx}
\usepackage{booktabs}
\usepackage{array}
\usepackage{amsfonts}
\usepackage{multirow}
\usepackage{dashrule}

\usepackage{amsmath}

\usepackage{tcolorbox}
\definecolor{xm-purple}{RGB}{210, 210, 210}
\definecolor{xm-grey}{RGB}{242,242,242}
\newtcolorbox[list inside=prompt,auto counter]{prompt}[1][]{
    colbacktitle=xm-purple!90,
    colback =xm-grey!30,
    coltitle=black,
    fontupper=\footnotesize,
    boxsep=5pt,
    left=0pt,
    right=0pt,
    top=0pt,
    bottom=0pt,
    boxrule=0.5pt,
    #1,
}

\title{Plan-over-Graph: Towards Parallelable LLM Agent Schedule}

\author{
Shiqi Zhang\textsuperscript{\rm 1, \# \thanks{\# Equal contribution.}},
Xinbei Ma\textsuperscript{\rm 1, \#},
Zouying Cao\textsuperscript{\rm 1},
Zhuosheng Zhang\textsuperscript{\rm 1},
Hai Zhao\textsuperscript{\rm 1, \dag }\\
  $^1$Shanghai Jiao Tong University \\
  \texttt{zsq259@sjtu.edu.cn, sjtumaxb@sjtu.edu.cn, zhaohai@cs.sjtu.edu.cn,}
  }
  
\makeatletter
\def\thanks#1{\protected@xdef\@thanks{\@thanks
        \protect\footnotetext{#1}}}
\makeatother

\begin{document}
\maketitle
\begin{abstract}
Large Language Models (LLMs) have demonstrated exceptional abilities in reasoning for task planning. However, challenges remain under-explored for parallel schedules. This paper introduces a novel paradigm, \textit{plan-over-graph}, in which the model first decomposes a real-life textual task into executable subtasks and constructs an abstract task graph. The model then understands this task graph as input and generates a plan for parallel execution. 
To enhance the planning capability of complex, scalable graphs, we design an automated and controllable pipeline to generate synthetic graphs and propose a two-stage training scheme. Experimental results show that our \textit{plan-over-graph} method significantly improves task performance on both API-based LLMs and trainable open-sourced LLMs. By normalizing complex tasks as graphs, our method naturally supports parallel execution, demonstrating global efficiency.
The code and data are available at \url{https://github.com/zsq259/Plan-over-Graph}.
\end{abstract}

\section{Introduction}
The commendable progress in large language models \citep{openai2023gpt4,templeton2024scaling, qwen2.5} has facilitated the impressive capability of agents for complicated, interactive tasks \cite{yao2023react, yao2022webshop, xi2024agentgym, ma-etal-2024-coco, yang2024swe}.
Recently studies have demonstrated that generating a plan before execution enhances agents' performance, referred to as the \textit{plan-then-execute} framework \cite{zhao2024large, hao2023reasoning, liu2023llm, zhang2023igniting, lin2024swiftsage}. 
Planning integrates global knowledge, enabling overall coherence rather than just local optimality \cite{qiao2024agent, ruan2024identifying}.
Planning breaks down complex tasks into subtasks as single-step operations, which is especially crucial for tasks requiring intricate workflows and precise action interfaces, such as UI control \cite{hong2023cogagent, wu2024oscopilot, zhang2024dynamic} and software engineering \cite{yang2024sweagent}. 
\begin{figure}[t]
\centering
\setlength{\belowcaptionskip}{-0.4cm}
\includegraphics[width=0.99\linewidth]{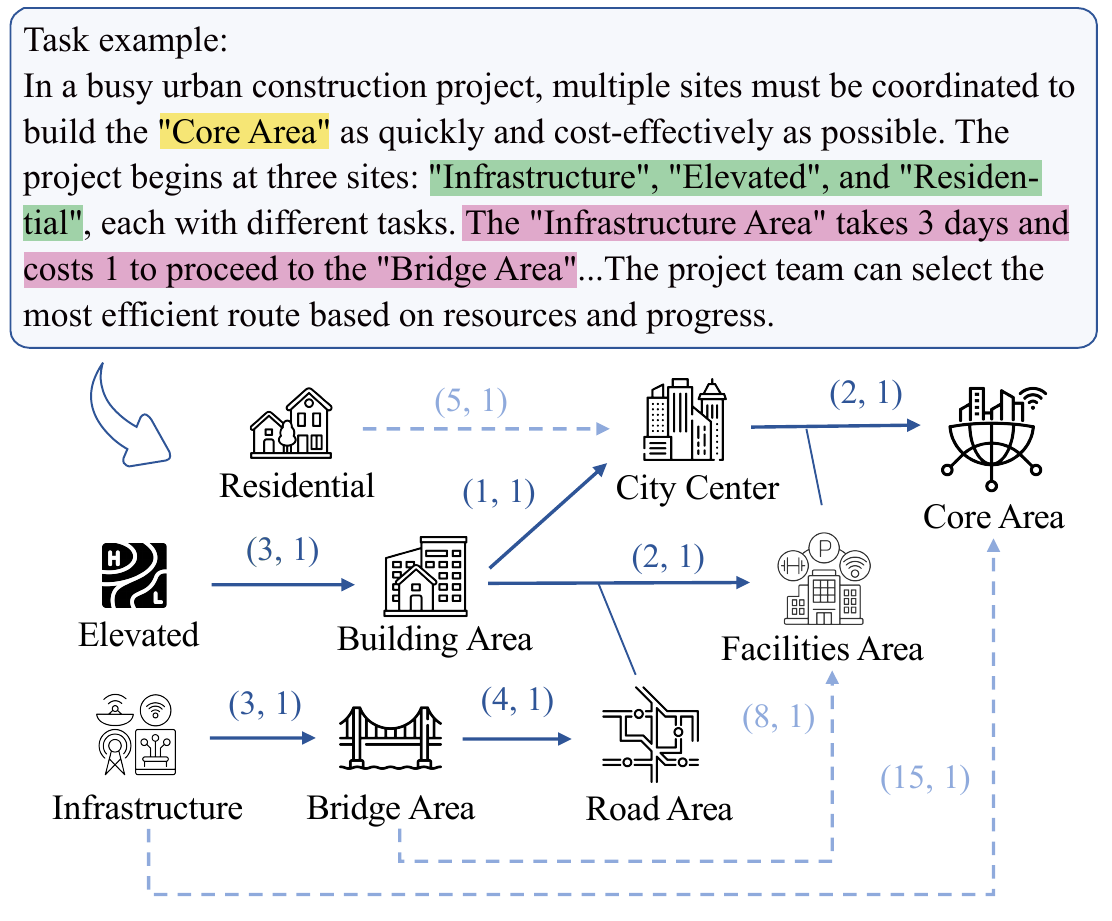}
\caption{An example of our task: from a realistic textual query to a parallel plan. The plan is represented as a graph. Edges are available rules, and \textit{Residential, Elevated, and Infrastructure} are the initial sources, \textit{Core} being the target. The solid edges denote the optimal plan under the constraint of time consumption.}
\label{fig:task_graph}
\end{figure}

Despite the inspiring progress, the parallelism of the plan remains under-explored. Multi-step agentic frameworks generally default to blocking pipelines, where each step waits until the previous ones to complete, regardless of whether it depends on their outcome \cite{wu2024oscopilot, gou2024critic}. 
The reasoning capabilities of agents are significantly stimulated by Chain-of-Thoughts (CoT) \cite{wei2023chainofthought}, enabling them to divide and conquer a complex task. Although the reasoning structure is expanded to trees and graphs \cite{yao2023tree, besta2024got, zhou2024language}, the actions for sub-tasks are taken sequentially.
However, these sub-tasks can run in parallel if independent of each other. While recent studies have explored the time efficiency of asynchronous execution, gaps remain when applied to real-world scenarios.

This paper dives into parallelism in planning for agents considering complex task graphs, as the example shown in \ref{fig:task_graph}.
We propose a new paradigm, \textit{plan-over-graph}, where the agent first explores rules and extracts a graph, then plans on the graphic structure under global consumption constraints.
For this novel method, we construct a dataset of complex tasks that involve parallel sub-tasks. Each sample is initialized with a connected directed acyclic graph, annotated with source and target nodes, along with feasible solutions and the optimal solution.
These graphs are further contextualized with appropriate scenarios by prompting an LLM to generate a textual description, finally forming realistic task descriptions in natural language.

We further improve our \textit{plan-over-graph} paradigm with a trainable scheme. 
As the scale of graphs expands, graph comprehension remains a challenge for agents \cite{Fatemi2023TalkLA, Chen2024LLaGALL, Luo2024GraphInstructEL, Dai2024HowDL}, leading to the performance bottleneck. 
Hence, we conduct a two-stage training strategy on abstract graphs.
During inference, the agent is prompted to extract textual queries to graphs and then plans over the graph with the trained adapter.
Our results are measured by comprehensive metrics including success rate, optimal accuracy, feasible accuracy, and efficiency.
Experiments achieve significant performance advancement on both API-base LLMs and open-sourced LLMs.
We further analyze the impact of graph structures scalability, demonstrate how parallel execution improves time efficiency, and identify common errors in task planning.

Our contributions can be summarized as follows:

$\circ$ We present \textit{plan-over-graph} for planning by enabling parallelism of sub-tasks, and construct a dataset of complex task graphs.

$\circ$ We enhance \textit{plan-over-graph} by training on task graphs and achieve significant improvement across LLMs.

$\circ$ We analyze that our approach achieves planning efficiency and maintains robustness across both diverse models and graph topology.

\section{Related Work}
This section introduces the background of agent planning and graph understanding of LLMs.

\subsection{Planning for LLM-based Agents}
Autonomous agents that interact with an environment to solve complex tasks \cite{yao2022webshop,fan2022minedojo}. Planning involves developing an action sequence before execution, leveraging global knowledge of the task and environment to suggest a logically consistent trajectory \cite{huang2024understanding}. 

Planning decomposes a complex task and selects a feasible trajectory based on global knowledge \cite{valmeekam2023on, wang2023plan, wu2024oscopilot}.
Searching strategies are applied to explore the optimal plan, such as depth-/breadth-first search and Monte Carlo tree search \cite{yao2023tree, zhou2024language,qi2024mutual, zhao2024large}.
When environmental feedback supplements perception heavily and updates the task knowledge, planning involves reflection to refine the trajectory incrementally \cite{shinn2024reflexion}.
A world model or a reward model integrates global knowledge and predicts the environment states or estimates rewards \cite{hao-etal-2023-reasoning, qiao2024agent}.

\subsection{LLM on Graphs}
As realistic intricate challenges can be formed as graphs, recent studies explore the LLMs' capabilities for reasoning with graphs.
Graph-of-Thought \cite{besta2024got, Ning2024DGoTDG} first proposes to transform the problem thinking into an arbitrary graph to enable the generation, aggregation, and refining of sub-tasks.
\citet{yao-etal-2024-got} also extract deductive triplets from contexts and build graphs.
Knowledge graphs support faithfulness and inference transparency for knowledge-intensive tasks \cite{Luo2023ReasoningOG, sun2024thinkongraph, Wen2023MindMapKG}. \citet{lin2024graphenhanced} combines graphs with natural language prompts for reasoning about asynchronous plans in real-life tasks, instructing model to either reason based on a given graph or to generate a graph themselves and then reason about it.




However, it is demonstrated that the capabilities of graph reasoning and understanding decrease as the scale and complexity of graphs increase. 
Empirical studies have observed a ``comprehension collapse'' phenomenon as the graph size increases \cite{Sui2023TableML, Cao2024GraphInsightUI}.
DARG \cite{zhang2024dargdynamicevaluationlarge} evaluates LLMs' reasoning capability on graphs and also reports a performance decrease with increasing complexity of graphs. 


Different from existing work, our paper further provides a more formal and scalable definition of the planning task's graph structure, which captures the inherent complexities and dependencies of the task. Our \textit{planning-over-graph} offers a general framework, independent of the specific nature of the task. Additionally, we demonstrate the effectiveness of this approach by training models on these graphs, achieving significant improvements in performance.

\section{Preliminary: Problem Statement}

This section formalizes the problem of planning on task graphs and presents a preliminary analysis to identify the key challenges.

\begin{figure*}[t]
\centering
\setlength{\belowcaptionskip}{-0.5cm}
\includegraphics[width=1\linewidth]{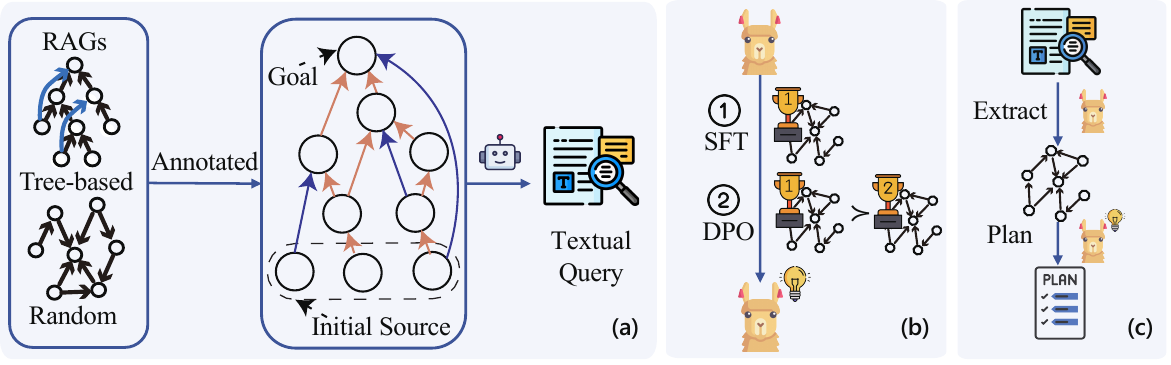}
\caption{The overview of our framework. (a) shows the data synthesis pipeline; (b) shows our training process; (c) displays the plan-over-graph paradigm.}
\label{fig:graph_pipeline}
\end{figure*}

\subsection{Formulation of Planning}
\label{sec:formulation_of_planning}

\textit{Planning} requires an agent to decompose a high-level task description into executable subtasks, schedule their execution under dependencies, optimize for time, cost, or multi-objective criteria, and finally achieve the goal. Formally, given a task description, the model generates a plan $P$ by solving the tuple $\langle G, \Omega \rangle$, where $G$ denotes the complex task and $\Omega$ denotes the global criteria.

Any high-level task description can be represented as a Directed Acyclic Graph (DAG), 
\begin{equation}
\setlength{\abovedisplayskip}{5pt}
\setlength{\belowdisplayskip}{5pt}
\begin{split}
&G = (T, E),\quad T = \{ t_1, t_2, \dots, t_n \},
\end{split}
\label{eq1}
\end{equation}

\noindent whose vertices are constrained by the precedence relationship formed by edges. e,g., $t_i \prec t_j$ means $t_i$ must precede $t_j$. 
One feasible plan $P$ is a subgraph of $G$ that satisfies $G$, $P \subset G$.



In realistic scenarios, there are criteria for constraints like execution time. We formally define a global function $\Omega$, where the optimal plan minimizes $\Omega$ while achieving the task,
\begin{equation}
\setlength{\abovedisplayskip}{5pt}
\setlength{\belowdisplayskip}{5pt}
\begin{split}
&P_{opt} = \arg \min\limits_{P} \Omega.
\end{split}
\label{eq2}
\end{equation}

\textbf{The measurements} examine whether the predicted plan is optimal or at least feasible, and also compute the metrics of global criteria.
Specifically, given each predicted plan $\hat{P}$ on a test dataset $D$, three kinds of metrics to evaluate its helpfulness:
(i) Optimal Rate: The proportion of optimal plans,
\begin{equation}
\setlength{\abovedisplayskip}{5pt}
\setlength{\belowdisplayskip}{5pt}
\begin{split}
&OR = |\hat{P} = P_{opt}|/|D|.
\end{split}
\label{eq3}
\end{equation}
(ii) Success Rate: The proportion of plans that successfully achieve the goal. 
\begin{equation}
\setlength{\abovedisplayskip}{5pt}
\setlength{\belowdisplayskip}{5pt}
\begin{split}
&SR = |\hat{P} \in \{P\}|/|D|.
\end{split}
\label{eq4}
\end{equation}
(iii) The values of considered global criteria $\Omega$.


\subsection{Challenges of Planning}
\label{sec:challenge}

Existing explorations leave two critical limitations. (i) Understanding graphs is currently a bottleneck in complex task planning for LLMs. \citet{lin2024graphenhanced} has shown that even with explicit graph representations, there is still a huge gap in handling complex graph topologies, suggesting unresolved challenges in structural reasoning. (ii) The scale of the currently considered graph is still very limited. WorFBench \cite{qiao2025benchmarking} considered graphs with majority nodes in the range of 2 to 10 steps. AsyncHow \cite{lin2024graphenhanced} most of the graph complexity $|V|+|E|$ are also between 10 and 20.

Considering these limitations, we design an experiment for a pilot study. We construct 100 random graphs with 10, 30, 50 nodes and ask LLM to find the shortest path as the solution. 
Table \ref{tab:preliminary_experiment} shows the accuracy of feasible paths and optimal paths, where the performance decreases sharply as the node number increases. Especially, and the optimal rate of 50-node graphs is only 6\%.

\begin{table}[htbp]
\setlength{\belowcaptionskip}{-0.2cm}
\setlength{\abovecaptionskip}{0.2cm}
  \centering
  \resizebox{0.48\textwidth}{!}{
  \setlength{\tabcolsep}{2.2mm}
    \begin{tabular}{l|cc}
      \toprule
      \textbf{Node Count} & \textbf{Optimal Rate$\uparrow$} & \textbf{Success Rate$\uparrow$} \\
      \midrule
      10 & 29.0 & 79.0 \\
      30 & 16.0 & 35.0 \\
      50 & 6.0 & 10.0 \\
      \bottomrule
    \end{tabular}
  }
  \vspace{-1mm}
  \caption{Llama-3.1-8B-Instruct on random graphs.}
  \label{tab:preliminary_experiment}
\end{table}



This empirical evidence demonstrates that the core bottleneck lies in planning competence on graph topology under complex constraints.
These findings inspire us to prioritize understanding capabilities on complex graphs with constraints to enhance the planning task.

\section{Methodology: Plan-over-Graph}

We propose the \textit{plan-over-graph} paradigm. Given a textual query, the LLM is prompted to gather information and build the task graph. Then, the task graph is input, on which we have the model perform planning. 
Then, we focus on enhancing the graph understanding for the \textit{plan} stage.

In the following Section \ref{sec:abstract_task_formulation} and \ref{sec:data_simulation}, we propose a data construction method to acquire a large amount of controllable graph data automatically.
Then, we design a training pipeline with these graph data (Section \ref{sec:graph_training}).
Finally, we combine these two steps to form the \textit{plan-over-graph} paradigm for inference. Our overall framework is shown is Figure \ref{fig:graph_pipeline}.



\subsection{Abstract Task Formulation}
\label{sec:abstract_task_formulation}
First, we redefine the planning task on graph structure.
Without the loss of generality, we consider time and cost limits for $\Omega$, the plan needs to minimize makespan or total cost.

\textbf{Task Graph.}
We define \textit{rules}, $R=\{r_i\}$, that allow parallel execution on a graph. For each node 
\begin{equation}
\setlength{\abovedisplayskip}{5pt}
\setlength{\belowdisplayskip}{5pt}
\begin{split}
&r = (S, t, \tau, c).
\end{split}
\end{equation}

A rule states that after 
$S$ is satisfied, 
$t$ can be executed with the required time $\tau$ and cost $c$. 

\textbf{Query.}
A query including prior knowledge and ultimate goal can be denoted as the initial node set and the target node as
\begin{equation}
\setlength{\abovedisplayskip}{5pt}
\setlength{\belowdisplayskip}{5pt}
\begin{split}
&q = (I, t_{\mathrm{target}}),
\end{split}
\end{equation}


\textbf{Plan.}
The overall plan $P$ is a set of several sub plans, each denoted as 
\begin{equation}
\setlength{\abovedisplayskip}{5pt}
\setlength{\belowdisplayskip}{5pt}
\begin{split}
&p = (S, t, D), \quad p_j \in D_i \iff t_j \in S_i
\end{split}
\end{equation}
where $S$ and $t$ are the preceding vertices and the subtask, determining a rule $r \in R$.
$D$ captures the dependencies between sub-plans. 


\textbf{Criteria}
Here, we define the two considered global criteria, time consumption and cost.
Subtasks that are independent of each other can be executed in parallel. In other words, $p_i$ only needs to wait for its dependency to end before starting execution. The end time of the execution of $p_i$ can be expressed as:
\begin{equation}
\setlength{\abovedisplayskip}{5pt}
\setlength{\belowdisplayskip}{5pt}
\begin{split}
&\text{End\_time}(p_i) = \max\limits_{p_j \in D_i } \left( \text{End\_time}(p_j) \right) + \tau_{p_i}, \\
&\Omega_{\text{time}}(P) = \max_{p \in P} \left( \text{End\_time}(p) \right).
\end{split}
\end{equation}

The global cost is defined as the sum of the cost values of all subtasks:
\begin{equation}
\setlength{\abovedisplayskip}{5pt}
\setlength{\belowdisplayskip}{5pt}
\begin{split}
&\Omega_{\text{cost}}(P) = \sum_{p \in P} c_p.
\end{split}
\end{equation}

We consider the time consumption as the main criterion. Hence, $P_{opt}$ is the value that minimizes
\begin{equation}
\setlength{\abovedisplayskip}{5pt}
\setlength{\belowdisplayskip}{5pt}
\begin{split}
\Omega(P) = \Omega_{\text{time}}(P) + \epsilon \cdot \Omega_{\text{cost}}(P)
\end{split}
\end{equation}
\noindent where $\epsilon \ll 1$.


According to our graph definition, the $\Omega$ is measured by Time Efficiency (TE) and Cost Efficiency (CE) as follows:
\begin{equation}
\begin{split}    
& \text{TE} = \frac{1}{n} \sum_{i=1}^{n} \frac{\Omega_{\text{time}}(P_i)}{\Omega_{\text{time}}(P_{opt})}, \\
& \text{CE} = \frac{1}{n} \sum_{i=1}^{n} \frac{\Omega_{\text{cost}}(P_i)}{\Omega_{\text{cost}}(P_{opt})}.
\end{split}
\end{equation}
The time and cost ratios of failed plans are assigned 4 as a penalty.

\subsection{Data Simulation}
\label{sec:data_simulation}
Following our definition above, we design an automated, controllable, and scalable pipeline to generate synthetic data, which consists of the following steps:

$\circ$ Generate a connected DAG.
Two distinct graph structures are employed: (i) Random DAGs.
(ii) To better conform to the hierarchical structure in reality and avoid a large number of shortcuts in random graph, we also construct a tree-based structure. We first construct a tree with a depth constraint of no more than 4, and then, a small number of ancestral and cross edges are added to introduce additional dependencies and enrich the graph structure. The edges are all directed to the root. The structural trade-offs between these two graph representations are analyzed in Section~\ref{sec:graph_features}. The graph is ensured to be connected, meaning there exists at least one path from the initial vertices to the target.

$\circ$ Define rules.
For each non-head subtask $t$ in the DAG, its predecessor vertices are randomly partitioned into groups as uniformly as possible. Then each predecessor group and $t$ forms the source and target of one rule. Each rule is assigned a random time value, which is sampled from a uniform distribution ranging from 1 to 50. The cost for all rules is fixed to 1, simplifying the cost structure while maintaining the focus on time optimization.

$\circ$ Define initial and target nodes. 
All head vertices
in the DAG are designated as the initial source vertices for the entire task. A target vertex is randomly selected from the tail vertices
, ensuring that the task has a well-defined goal.

$\circ$ Annotate optimal and feasible solutions. 
A dynamic programming algorithm is applied to compute the labels of each solution of graphs. The optimal solution is the gold label.


$\circ$ Construct query descriptions.
We then generate textual queries based on those graphs. We prompt an LLM to transform graphs into real-life scenario descriptions. 
To ensure consistency between the generated task and the original graph, we let the LLM perform the self-correction to verify that the query description matches the original graph.

\subsection{Training Scheme}
\label{sec:graph_training}

In this section, we focus on optimizing the model's ability on abstract graphs and improve its parallel planning capabilities. Our training has two stages: supervised fine-tuning and direct preference optimization.

\textbf{Supervised Fine-Tuning.}
We fine-tune an LLM on our abstract task datasets. This enables the model to solve planning tasks using graph representations of the problem space. We use the Low-Rank Adaptation (LoRA) method \cite{hu2022lora}, which allows efficient adaptation of large pre-trained models by learning a small-size adapter.
Two setups are considered for SFT: (i) fine-tuning with optimal data instances. (ii) to enable the model to learn both optimal and feasible solutions, we select the second-best solutions and mix them with the optimal solutions as the training data. 

\textbf{Direct Preference Optimization.}
Following fine-tuning, direct preference optimization (DPO) \cite{rafailov2023direct} is applied to distinguish the optimal solution from feasible ones. For each sample, the second-best solution works as the rejected output, while optimal solutions are the chosen output. This step further refines the model’s ability to prioritize optimal solutions over feasible ones.

After training, we aggregate the extracting and planning steps during the inference. Given a query with a goal description, we first extract the task graph from the description, then we generate the plan on the graph with the trained adapter loaded. 

\section{Experiment}
\subsection{Dataset}
\label{sec:dataset_setup}

\begin{table}[t]
\setlength{\belowcaptionskip}{-0.4cm}
\setlength{\abovecaptionskip}{0.2cm}
\small
\centering
\scalebox{0.975}{
\begin{tabular}{l|l|l|l|c} 
\toprule
\multirow{2}{*}{\textbf{Set}} & \multirow{2}{*}{\textbf{Nodes}} & \textbf{Graph} & \textbf{Edge} & \multirow{2}{*}{\textbf{Samples}}\\
& & \textbf{Structure} & \textbf{Relation} & \\
\midrule
\multirow{6}{*}{Training} 
& \multirow{2}{*}{10} & Random & Uniform & 2000 \\
& & Tree-based & Linear & 2000 \\
\cmidrule{2-5}
& \multirow{2}{*}{30} & Random & Uniform & 2000 \\
& & Tree-based & Linear & 2000 \\
\cmidrule{2-5}
& \multirow{2}{*}{50} & Random & Linear & 2000 \\
& & Tree-based & Linear & 2000 \\
\midrule
\multirow{12}{*}{Testing}
& \multirow{3}{*}{10} & Random & Linear & 100 \\
& & Tree-based & Linear & 100 \\
& & Random & Uniform & 1000 \\
\cmidrule{2-5}
& \multirow{2}{*}{20} & Random & Linear & 100 \\
& & Tree-based & Linear & 100 \\
\cmidrule{2-5}
& \multirow{3}{*}{30} & Random & Linear & 100 \\
& & Tree-based & Linear & 100 \\
& & Random & Uniform & 1000 \\
\cmidrule{2-5}
& \multirow{2}{*}{40} & Random & Linear & 100 \\
& & Tree-based & Linear & 100 \\
\cmidrule{2-5}
& \multirow{2}{*}{50} & Random & Linear & 100 \\
& & Tree-based & Linear & 100 \\
\bottomrule
\end{tabular}
}
\vspace{-1mm}
\caption{Detailed statistics of our training and testing dataset.}
\label{tab:dataset}
\end{table}

\textbf{Task Graph.}
The theoretical edge of a connected graph with node count $n$ range spans from $n-1$ to $n(n-1)/2$. Massive edges lead to excessively long input when $n$ is large. And tree-based graph structure also cannot have too many edges. Therefore, we adopt two practical strategies to avoid this: (i) linear scaling with edges $\in[2n,3n]$ for random graphs and $\in[n,1.5n]$ for tree-based graphs; (ii) uniform distribution across the full edge range, which is only used on random graphs.

The statistics of our synthetic data are shown in Table \ref{tab:dataset}. The training set contains 12,000 training instances, divided equally across three node scales (10, 30, 50 nodes) and random and tree-based DAG structures. It employs uniform distribution edge configuration for 10/30-nodes graphs but restricts 50-nodes graphs to linear scaling. We generate 1000 input instances for each node scale and graph structure. Each input corresponds to an optimal solution and a chosen feasible solution. The testing set comprises two components: (i) baseline tests with linear edge scaling across node counts (10, 20, 30, 40, 50 nodes), each node count and graph structure containing 100 instances; (ii) edge-variation tests specifically for 10/30-nodes random graphs with uniformly distributed edges, to evaluate the model's ability to understand graphs as the number of edges changes. Due to the wide range of changes in the number of edges, we generate 1,000 instances each.

\textbf{Textual Query.}
To systematically evaluate the parallel planning capacity of the model in real-world task scenarios and validate our \textit{plan-over-graph} paradigm, we construct an evaluation dataset utilizing the DeepSeek-R1 \cite{deepseekai2025deepseekr1incentivizingreasoningcapability} model. This dataset synthesizes 200 tasks derived from some real-world problem domains, where each task specification is transformed from graphs into executable workflow descriptions. The query data statistics are outlined in Appendix \ref{query_statistcs}.



\subsection{Baseline}

We evaluate our method against several baseline models, including API-based LLMs GPT-4o \cite{hurst2024gpt} and Claude 3.5 Sonnet \cite{claude} , and open-source LLMs, Llama-3.1-8B-Instruct \cite{dubey2024llama} and Qwen2.5-7B-Instruct \cite{qwen2.5}. These models are selected for their strong performance and wide use. The evaluation metrics are detailed in Section \ref{sec:abstract_task_formulation}. Detailed setups can be found in Appendix \ref{sec:training_setups}.

\subsection{Main Results}

Table \ref{tab:combined_results} presents experimental results. Our results answer the following three key questions.

\textit{Q1: Can our training method teach the LLMs plan on the graph? Which method derives the best performance?}
The overall results for each model are shown in the upper part of Table~\ref{tab:combined_results}. 
Overall, it can be observed that \textbf{training largely improves the planning performance and the two-stage training achieves even better scores.}
Without training, Claude demonstrates high success rates (90.0\%) but relatively lower optimal rates (39.2\%). GPT-4o and Llama have similar success rates (51.3\% and 52.3\%), but GPT-4o achieves a higher Optimal Rate (14.1\%) than 1.8\% of Llama. Qwen shows weaker performance in both success and optimal rates (13.2\% and 0.5\%).

Only training on optimal solutions significantly improves graph understanding and planning ability, leading to a 75.7\% success rate and a 61.4\% optimal rate for Llama. Mixing feasible solutions further improves the performance to 86.1\% and 67.5\%. The best results of optimal rates (71.6\%) are derived from the two-stage training, combining SFT on mixed data and DPO, maintaining high success rates (83.6\%). This is because the model further tends to choose the optimal solution through DPO. We also observed consistent performance improvements on Qwen, which is inferior to Llama. 

\begin{table*}[htbp]
\setlength{\belowcaptionskip}{-0.4cm}
\setlength{\abovecaptionskip}{0.2cm}
  \centering
  \small
  \resizebox{\textwidth}{!}{
    \begin{tabular}{llllll}
      \toprule
      \multicolumn{6}{c}{\textbf{Overall Results}} \\
      \midrule
      \textbf{Model} & \textbf{Optimal Rate$\uparrow$} & \textbf{Success Rate$\uparrow$} & \textbf{Feasible Rate} & \textbf{Avg Time Ratio$\downarrow$} & \textbf{Avg Cost Ratio$\downarrow$} \\
      \midrule
      Claude 3.5 Sonnet& 39.2 & \textbf{90.0} & 50.8 & 1.545 & 1.589 \\
      GPT-4o & 14.1 & 51.3 & 37.2 & 2.657 & 2.889 \\
      \multicolumn{6}{c}{\vspace{-3.2mm}\hdashrule[1.5ex]{16.8cm}{0.5pt}{4pt}} \\
      Llama-3.1-8B-Instruct & 1.8 & 52.3 & 50.5 & 2.616 & 4.512 \\
      Llama-3.1-8B-Instruct\textsubscript{\texttt{opt} SFT} & 
      61.4\scriptsize{+59.6} & 
      75.7\scriptsize{+23.4} & 14.3 & 1.746\scriptsize{-0.870} & 1.769\scriptsize{-2.743} \\
      Llama-3.1-8B-Instruct\textsubscript{\texttt{opt+feas} SFT} & 67.5\scriptsize{+65.7} & 86.1\scriptsize{+33.8} & 18.6 & 1.507\scriptsize{-1.109} & \textbf{1.498}\scriptsize{-3.014} \\
      Llama-3.1-8B-Instruct\textsubscript{\texttt{opt+feas} SFT + \texttt{opt} DPO} & 
      \textbf{71.6}\scriptsize{+69.8} & 
      83.6\scriptsize{+31.3} & 12.0 & \textbf{1.502}\scriptsize{-1.114} & 1.520\scriptsize{-2.992} \\
      \multicolumn{6}{c}{\vspace{-3.2mm}\hdashrule[1.5ex]{16.8cm}{0.5pt}{4pt}} \\
      Qwen2.5-7B-Instruct & 0.5 & 13.2 & 12.7 & 3.612 & 4.072 \\
      Qwen2.5-7B-Instruct\textsubscript{\texttt{opt+feas} SFT + \texttt{opt} DPO} & 
      27.0\scriptsize{+26.5} &
      75.8\scriptsize{+62.6} & 48.8 & 2.191\scriptsize{-1.421} & 2.029\scriptsize{-2.043} \\
      \midrule[1.5pt]
      \multicolumn{6}{c}{\textbf{Textual Query Results}} \\
      \midrule
      \textbf{Method} & \textbf{Optimal Rate$\uparrow$} & \textbf{Success Rate$\uparrow$} & \textbf{Feasible Rate} & \textbf{Avg Time Ratio$\downarrow$} & \textbf{Avg Cost Ratio$\downarrow$} \\
      \midrule
      Claude Plan & 14.5 & 89.5 & 75.0 & 1.904 & 2.302 \\
      Claude Extract + Plan & 
      41.5\scriptsize{+27.0} & 
      \textbf{93.5}\scriptsize{+4.0} & 52.0 & \textbf{1.514}\scriptsize{-0.390} & 1.689\scriptsize{-0.613} \\
      \multicolumn{6}{c}{\vspace{-3.2mm}\hdashrule[1.5ex]{16.8cm}{0.5pt}{4pt}} \\
      Llama Plan & 0.0 & 19.0 & 19.0 & 3.433 & 4.103 \\
      Llama Extract + Plan & 
      3.5\scriptsize{+3.5} & 
      38.0\scriptsize{+19.0} & 34.5 & 2.952\scriptsize{-0.481} & 3.553\scriptsize{-0.550} \\
      Llama Extract + Llama-trained Plan & 
      \textbf{72.5}\scriptsize{+72.5} & 
      83.0\scriptsize{+64.0} & 17.0 & 1.540\scriptsize{-1.893} & \textbf{1.526}\scriptsize{-2.577} \\
      \bottomrule
    \end{tabular}
  }
  \vspace{-1mm}
  \caption{Experiment results: the upper part shows results on all baseline test sets; the lower part shows results on real-life tasks}
  \label{tab:combined_results}
\end{table*}

\begin{figure*}[t]
\setlength{\belowcaptionskip}{-0.2cm}
\setlength{\abovecaptionskip}{0.2cm}
\centering
\includegraphics[width=1\linewidth]{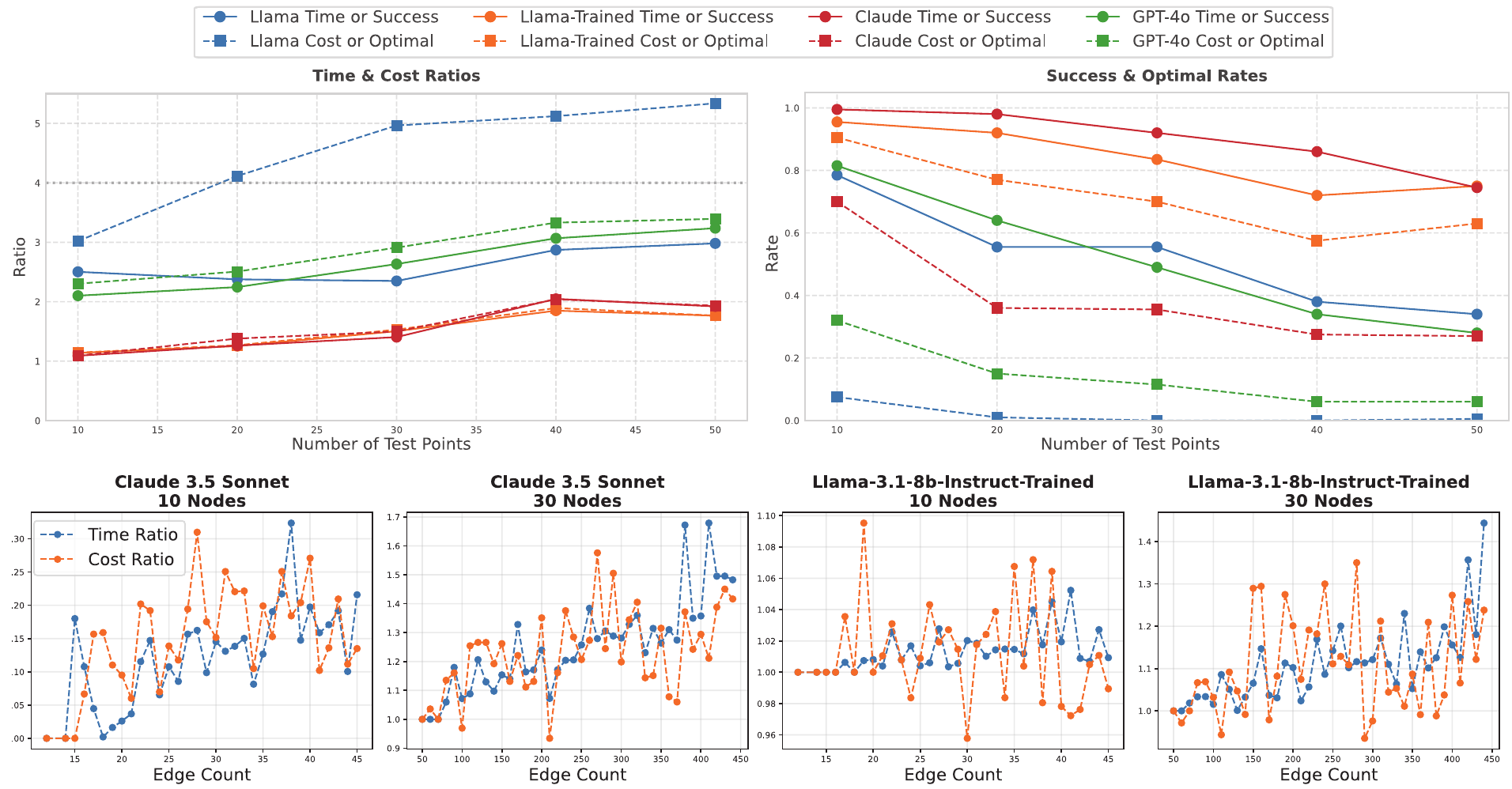}
\vspace{-1mm}
\caption{The upper part of this figure shows model performance across different node counts. The left plot shows time and cost ratio change with the number of points, and the right shows success and optimal rate. The lower part shows Claude and our trained Llama average time and cost ratio across different edge counts.}
\label{fig:combined_results}
\end{figure*}

\textit{Q2: What phenomena does the expansion of the graph scale lead to?}
As shown in the upprt part of Figure \ref{fig:combined_results}, when the number of nodes increases, which leads to a larger graph structure, the overall performance of all models decreases. For time and cost ratio, these models have shown similar sensitivity to the node count. However, the cost ratio of Llama is quite obviously increased, which shows its tendency to select more subtasks on larger graphs. For success rate, GPT-4o and Llama drop sensitively. When the number of nodes reaches 30, GPT-4o even falls below Llama. However, Claude and our trained model continue to demonstrate strong capabilities with less sensitive drops. Claude still suffers between 10 and 20 points on the optimal rate, indicating a gap in the understanding of larger-scale graphs. Across all node counts, our trained Llama significantly outperforms all other models on the optimal rate.

The lower part of Figure \ref{fig:combined_results} shows the results of Claude and our trained Llama on the full range of edge counts for 1000 cases with 10/30-nodes, respectively. For 10 nodes, due to the number of nodes being small, both models have demonstrated robustness to changes in the number of edges across the metrics. For 30 nodes, as the number of edges increases, the difficulty for the model to find the optimal solution increases more significantly. Therefore, the average time ratio for both is on the rise. The average cost ratio increases less significantly, because as the graph becomes denser, the number of feasible solutions also increases, allowing the model to complete tasks by selecting fewer subtasks, though not in the most optimal time.

\textit{Q3: Can our plan-over-graph method improve the planning performance on textual queries?}
The lower part of Table \ref{tab:combined_results} presents the results of real-life queries, \textbf{showing that our plan-over-graph method consistently improves different models' performance.} 
When planning without extraction, Claude achieves a 89.5\% success rate but struggles with the 14.5\% optimal rate and a high 1.904 time ratio. Llama achieves a success rate of 19\%, with no optimal plan available, causing a 3.433 time ratio. With our plan-over-graph framework, the optimal rate of Claude improves to 41.5\%, and the success rate of Llama also improves to 38\%, with both time ratios decreased. The model trained for planning shows greatly improved performance, surpassing even Claude in 72.5\% optimal rate with a 83\% success rate.

In summary, these results demonstrate that: First, the plan-over-graph method improved model performance. Second, training on the plan significantly enhanced model performance.

\section{Analysis}

In this section, we discuss our dataset and the detailed results of the experiment.

\subsection{Graph Features}
\label{sec:graph_features}
This section discusses (i) our considerations regarding the graph structure; (ii) the impact of changes in the number of nodes and edges in the graph on the planning capability of the model.

\textbf{Graph Structures.}
The tree-based structure is designed to better reflect real-world parallel scenarios, offering a stronger hierarchical organization that facilitates the generation of more reasonable specific scenarios. To ensure a clear distinction between parallel and non-parallel execution, we implement the following strategies: (i) controlling the depth of the tree to manage the number of branches, and (ii) grouping nodes such that the number of nodes in each group does not exceed two-thirds of the total number of predecessor nodes. In addition, we also used an undefined random graph structure to verify the robustness of the model.


\textbf{Impact of Node and Edge Counts}
We calculated the absolute value of correlation coefficients and slopes of normalized four metrics with changes in the number of points and edges. 
Overall, the impact of the edge count is smaller than the node count.
For node counts, almost all metrics of the models showed strong correlation coefficients between 0.8 and 1.0. 
For edge variations on 10 nodes, both model shows low correlation coefficients which are less than 0.5, indicating robustness to changes in the number of edges on graphs with fewer points. 
However, on 30 nodes, Claude has higher correlation coefficients on all four metrics than our trained Llama, which are more than 0.7 correlation coefficients, demonstrating lower stability. Please refer to Appendix \ref{sec:supplementary_of_the_experiments} for details.

\subsection{Time Efficiency}

Our tasks inherently support parallel execution of subtasks, yet most existing methods do not consider parallelism during planning, leading to unnecessary waiting times. 

We calculate the time ratio of parallel execution to sequential execution (that is, the sum of all subtask durations) in the plans. Table \ref{tab:parallel_sequential_ratio} shows the results of optimal labels, and outputs of our trained Llama and Qwen.
The results demonstrate that the capability of planning parallel solutions can significantly reduce time compared to blocking sequential execution. 
Such efficiency is more significant as the graph scales.
Specifically, plans from Llama and Qwen show high cost ratios, indicating that there are many redundant subtasks. When executing these plans sequentially, the inefficiency will be further amplified, leading to a low ratio of the parallel execution time to the sequential.

\begin{table}[t]
\setlength{\belowcaptionskip}{-0.4cm}
\setlength{\abovecaptionskip}{0.2cm}
    \centering
    \scalebox{0.76}{
    \begin{tabular}{@{}l| cc cc cc cc cc cc cc@{}}
    \hline
     & \multicolumn{2}{c}{\textbf{Optimal}}
     & \multicolumn{2}{c}{\textbf{Llama-trained}} 
     & \multicolumn{2}{c}{\textbf{Qwen-trained}}
     \\ \cmidrule(l){2-7} 
    \multirow{-2}{*}{\textbf{Node Count}} 
    & R & \multicolumn{1}{c|}{T}
    & R & \multicolumn{1}{c|}{T} 
    & R & T \\
    \hline
    
    10
    & 0.88 & \multicolumn{1}{c|}{0.92} 
    & 0.88 & \multicolumn{1}{c|}{0.92}
    & 0.88 & 0.93 \\
    
    20
    & 0.76 & \multicolumn{1}{c|}{0.74}
    & 0.77 & \multicolumn{1}{c|}{0.80} 
    & 0.79 & 0.79 \\
    
    30
    & 0.74 & \multicolumn{1}{c|}{0.75}
    & 0.75 & \multicolumn{1}{c|}{0.75}
    & 0.73 & 0.68 \\
    
    40
    & 0.70 & \multicolumn{1}{c|}{0.68}
    & 0.71 & \multicolumn{1}{c|}{0.68}
    & 0.68 & 0.61 \\
    
    50
    & 0.68 & \multicolumn{1}{c|}{0.62}
    & 0.73 & \multicolumn{1}{c|}{0.61}
    & 0.70 & 0.56 \\
    \hline
    \end{tabular}
    }
    \vspace{-1mm}
    \caption{The ratio of the parallel execution time of the plans provided by each model at the test cases to the sequential execution time. R represents the random graph structure, and T represents the tree-based structure.}
    \label{tab:parallel_sequential_ratio}
\end{table}

\subsection{Wrong Case Study}
\textbf{Task Graph.}
The wrong cases on abstract graphs fall into two types.
(i) \textit{Invalid subtask}, where the plan includes subtasks without corresponding transformation rules, and (ii) \textit{Unavailable source}, where the required source for a subtask is not achieved during execution. The latter indicates either a failure to consider the source availability during planning or an incorrect handling of dependencies. Table \ref{tab:wrong_type_proportion} shows the proportion of two error types. After training, the source dependency has almost been resolved, but the hallucination of invalid subtasks is currently the performance bottleneck.

\begin{table}[htbp]
\setlength{\belowcaptionskip}{-0.2cm}
\setlength{\abovecaptionskip}{0.2cm}
  \centering
  \small
    \begin{tabular}{l|cc}
      \hline
      \multirow{2}{*}{\textbf{Model}} & \textbf{Invalid} & \textbf{Unavailable} \\ & \textbf{Subtask} & \textbf{Source} \\
      \hline
      Claude 3.5 Sonnet & 0.4 & 9.6 \\
      GPT-4o & 4.7 & 44.0 \\
      Llama-3.1-8B-Instruct & 17.6 & 30.1 \\
      Llama-3.1-8B-Instruct-Trained & 11.6 & 4.8 \\
      Qwen2.5-7B-Instruct & 60.7 & 26.1 \\
      Qwen2.5-7B-Instruct-Trained & 19.9 & 4.3 \\
      \hline
    \end{tabular}
  \vspace{-1mm}
  \caption{The proportion of two error causes in all test cases of the model.}
  \label{tab:wrong_type_proportion}
\end{table}

\noindent\textbf{Textual Query.}
\label{sec:wrong_case_in_specific_task}
Failure to extract essential rules for subtasks will compromise the model's overall performance.
However, interestingly, even with extraction errors, the model can still complete the task correctly if the subsequent planning does not encounter incorrect rules. Results of Llama show that the extraction step significantly improves the baseline success rate, and while the trained model's success rate is slightly lower than that of Claude, its optimal rate is superior. After taking a closer look, finding that only 15\% matched the original graph exactly. However, the average similarity for mismatched cases was 82\%, indicating minimal impact. This supports our focus on improving the model's planning capabilities on abstract graphs.


\section{Conclusion}

We present plan-over-graph, a novel paradigm to enhance parallelism in LLM-based agentic planning. Our approach extracts task dependencies as structured graphs, then optimizes parallel planning through graph-aware reasoning. We develop a synthetic dataset annotated with directed acyclic graphs and propose a two-stage training scheme. Experimental results demonstrate significant improvements.
Our analysis further reveals that the graph structure inversely affects model performance and the time reduction brought by parallelism. This work establishes a framework for parallel agentic systems, bridging the gap between abstract graphs and real-world applications.



\section*{Limitations}
We acknowledge the limitations of this work. (i) Although we believe and verify that the ability to plan on the graph is more important than extraction, open source models have also shown certain flaws in extraction. (ii) In reality, the model's plan can be a dynamic process that interacts with the environment, where the model can refine the previously given plan through perception. Our future work will focus on these two directions.

\bibliography{custom}

\appendix

\section{Prompt Template}
\label{sec:prompt_template}

We show the prompt templates following examples.

\begin{prompt}[title={Prompt template for graph planning}]
You are given a set of transformation rules, where each rule consists of source nodes (materials or subtasks), target nodes (resulting materials or tasks), the time required to complete the transformation, and a cost associated with the transformation. Your goal is to plan a path from the initial nodes to the target node, supporting parallel transformations, to obtain the target node in the shortest time possible, while minimizing the total cost.

Input format:

- Transformation rules: A list of dictionaries, where each dictionary represents a transformation rule and contains:

\ \ \ \   - source: A list of source nodes (the prerequisites for the transformation).

\ \ \ \   - target: A list of target nodes (the result of the transformation).

\ \ \ \   - time: The time required to complete the transformation (an integer).

\ \ \ \   - cost: The cost associated with the transformation (an integer).

- Initial nodes: A list of strings representing the available nodes at the start.

- Target node: A string representing the node that needs to be obtained.

Output format:

- Plan: A list of subtasks, where each subtask is a JSON object with the following fields:

\ \ \ \   - name: The name of the subtask or node being completed. The default name format is "Subtask" followed by a sequence number.
  
\ \ \ \   - source: A list of source nodes involved in this subtask. The sources must be products you already have or can obtain through previous steps.
  
\ \ \ \   - target: The target node resulting from this subtask. Both the source and target must conform to a given rule and cannot be assumed or self-created.
  
\ \ \ \   - dependencies: A list of dependencies (other subtask names) that need to be completed before this subtask can be executed. This ensures the execution order between subtasks, and the dependencies must provide the required sources for this subtask.

Important: 

- The generated JSON must strictly follow the JSON format. The following rules must be strictly adhered to:

\ \ \ \   - All keys and values must be enclosed in double quotes.

\ \ \ \   - All elements in arrays must be separated by commas.

\ \ \ \   - All fields in the JSON must be complete and correctly formatted, with no missing or incorrect elements.

- All planned steps must comply with a given rule.

- All substances involved must conform to the given rules.

Your task is to generate the final plan in the specified JSON format, minimizing both the completion time and total cost. Do not provide any implementation code.

Here is an example to better understand the task:

\{graph\_planning\_example\}

Now, based on the following transformation rules, initial nodes, and target node, please provide an optimal plan that allows the target node to be obtained in the shortest time with the minimal total cost, supporting parallel transformations.

Only include necessary steps that are required for the fastest completion with the least cost. Do not add any extra or redundant transformation steps.

Task:

```json

\{task\}

```

Your task is to generate the final plan in the specified JSON format. Do not provide any implementation code.
\end{prompt}

\begin{prompt}[title={Prompt template for query planning}]
For the input task, please provide an optimal plan that allows the target to be obtained. 
Minimize the cost under the premise of the shortest time.

Projects without dependencies can be completed in parallel to improve overall efficiency. 

Please provide the final solution in JSON format:

- Plan: A list of subtasks, where each subtask is a JSON object with the following fields:

\ \ \ \   - name: The name of the subtask or node being completed. The default name format is "Subtask" followed by a sequence number.

\ \ \ \   - source: A list of source nodes involved in this subtask. The sources must be products you already have or can obtain through previous steps.

\ \ \ \   - target: The target node resulting from this subtask. Both the source and target must conform to a given rule and cannot be assumed or self-created.

\ \ \ \   - dependencies: A list of dependencies (other subtask names) that need to be completed before this subtask can be executed. This ensures the execution order between subtasks, and the dependencies must provide the required sources for this subtask.
Here is an example:

Input:

\{query\_example\}

Output:

```json

\{query\_example\_plan\}

```

Input:

\{task\}

Output:

```
\end{prompt}

\begin{prompt}[title={Prompt template for extracting graph from query}]
Task: Extract structured transition rules from unstructured workflow narratives.  
Objective: Identify all transitions between nodes in the text. For each transition, extract:  

- Source nodes (prerequisites)  

- Target nodes (outcomes)  

- Time (duration)  

- Cost (numeric resource units)  

Additionally, determine the initial\_source (starting node) and target (final node).  

Input: A story describing a workflow process. Example phrases may include:  

- "From [NodeA], proceed to [NodeB] in X days at a cost of Y units"  

- "When both [NodeA] and [NodeB] are ready, [NodeC] can be completed in X days at a cost of Y units"

- Shortcuts like "directly from [NodeA] to [NodeC] in X days at a cost of Y units"  

Output: 

A JSON object with:  

1. "rules": A list of transition rules, each containing:  

   - "id" (sequential integer starting from 0)  
   
\ \ \ \    - "source" (list of node IDs, e.g., ["N1"])  

\ \ \ \    - "target" (list of node IDs, e.g., ["N2"])  

\ \ \ \    - "time" (numeric value)  

\ \ \ \    - "cost" (numeric value)  

2. "initial\_source": List of starting node IDs (e.g., ["N1"])  

3. "target": Final node ID (e.g., "N8")  

Your Task: Convert the following story into the JSON format above. Ensure:  

1. All transitions are captured, including multi-source dependencies and shortcuts  

2. Node IDs (e.g., N1, N2) are preserved exactly as written  

3. Time and cost values are strictly numeric  

4. Follow the JSON schema precisely  

Example Input Story:  

\{query\_example\}

Example Output:

```json

\{query\_example\_plan\}

```

Input Story:

\{task\}

Output:
\end{prompt}

\begin{prompt}[title={Prompt template for generating query from graph}]
Transform this abstract task into a specific task in a real-world scenario, noting the following:

1. Tasks without dependencies can be executed in parallel.

2. Please express the instructions in complete natural language without explicitly listing the rules.

3. As long as there is one path that reaches the final goal, it is considered successful.

4. The source of a rule must be fully achieved before proceeding with the rule and obtaining its target.

5. You must strictly follow the rules I have given, make sure the rules in your story correspond one-to-one with the rules I have provided, and the sum of rules in your story must be equal to the sum of rules in the task.

6. You must explicitly mention both the time and cost associated with each rule in the story.

7. You only need to write the rules as a story, without offering any additional evaluation comments or introductory remarks.

Here is an example from another task for reference:

Input:
```json

\{query\_example\_plan\}

```

OutPut:

\{query\_example\}

Input:

```json

\{task\}

```

Output:
\end{prompt}

\begin{prompt}[title={Examples}]
\label{sec:examples}
\color{orange}{graph\_planning\_example:}
\color{black}{}

Task:

```json

\{
    "rules": [
        \{
            "source": ["N1"],
            "target": ["N2"],
            "time": 3,
            "cost": 1
        \},
        \{
            "source": ["N6"],
            "target": ["N3"],
            "time": 4,
            "cost": 1
        \},
        \{
            "source": ["N2", "N3"],
            "target": ["N4"],
            "time": 2,
            "cost": 1
        \},
        \{
            "source": ["N4"],
            "target": ["N5"],
            "time": 1,
            "cost": 1
        \},
        \{
            "source": ["N2"],
            "target": ["N5"],
            "time": 5,
            cost": 1
        \}
    ],
    "initial\_source": ["N1", "N6"],
    "target": "N5"
\}

```

Expected output:

```json

[
    \{
      "name": "Subtask1",
      "source": ["N1"],
      "target": ["N2"],
      "dependencies": []
    \},
    \{
      "name": "Subtask2",
      "source": ["N6"],
      "target": ["N3"],
      "dependencies": []
    \},
    \{
      "name": "Subtask3",
      "source": ["N2", "N3"],
      "target": ["N4"],
      "dependencies": ["Subtask1", "Subtask2"]
    \},
    \{
      "name": "Subtask4",
      "source": ["N4"],
      "target": ["N5"],
      "dependencies": ["Subtask3"]
    \}
]

\color{orange}{query\_example:}
\color{black}{}

In a busy urban construction project, multiple sites must be coordinated to build the "Core Area(N9)" as quickly and cost-effectively as possible. The project begins at three sites: "Infrastructure(N1)," "Elevated(N3)," and "Residential(N7)," each with different tasks. The "Infrastructure Area(N1)" takes 3 days and costs 1 to proceed to the "Bridge Area(N2)", while the "Elevated Area(N3)" moves to the "Building Area(N4)" in 3 days and at a cost of 1. The "Bridge Area(N2)" connects to the "Road Area(N5)" in 4 days and costs 1, and can directly connect to the "Facilities Area(N6)" in 8 days at a cost of 1. The "Building Area(N4)" partners with the "Road Area(N5)" to build the "Facilities Area(N6)" in 2 days and at a cost of 1. The "Residential Area(N7)" takes 5 days and costs 1 to reach the "City Center Area(N8)", while the "Building Area(N4)" directly reaches it in 1 day and costs 1. Once the "Facilities(N6)" and "City Center(N8)" areas are ready, they combine to complete the "Core Area(N9)" in 2 days at a cost of 1. The "Infrastructure Area(N1)" has a shortcut to bypass other areas and reach the "Core Area(N9)" in 15 days at a cost of 1. The project team can select the most efficient route based on resources and progress.

\color{orange}{query\_example\_plan:}
\color{black}{}
\{"rules": [\{ 'id': 0, "source": ["N1"], "target": ["N2"], "time": 3, "cost": 1 \}, \{ 'id': 1, "source": ["N3"], "target": ["N4"], "time": 3, "cost": 1 \}, \{ 'id': 2, "source": ["N2"], "target": ["N5"], "time": 4, "cost": 1 \}, \{ 'id': 3, "source": ["N4", "N5"], "target": ["N6"], "time": 2, "cost": 1 \}, \{ 'id': 4, "source": ["N2"], "target": ["N6"], "time": 8, "cost": 1 \}, \{ 'id': 5, "source": ["N7"], "target": ["N8"], "time": 5, "cost": 1 \}, \{ 'id': 6, "source": ["N4"], "target": ["N8"], "time": 1, "cost": 1 \}, \{ 'id': 7, "source": ["N6", "N8"], "target": ["N9"], "time": 2, "cost": 1 \}, \{ 'id': 8, "source": ["N1"], "target": ["N9"], "time": 15, "cost": 1 \}, ], "initial\_source": ["N1", "N3", "N7"], "target": "N9"\}

```
\end{prompt}

\section{Textual Query Statistics}
\label{query_statistcs}

\begin{figure}[ht]
\centering
\includegraphics[width=1\linewidth]{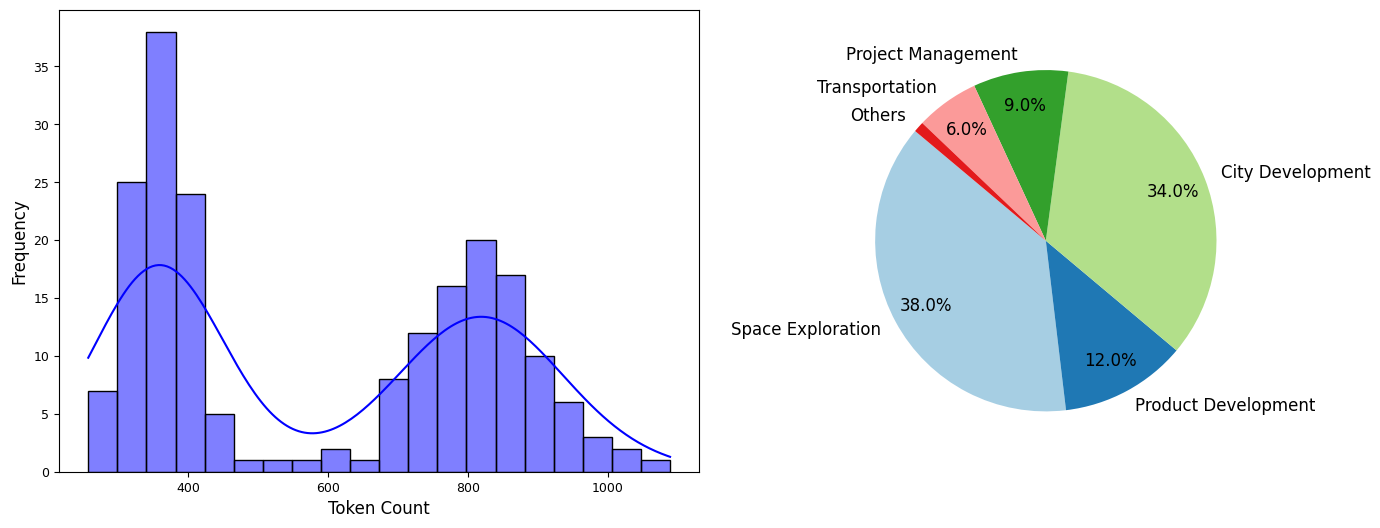}
\caption{Statistics on our synthetic query. The bar chart on the left displays the distribution of tokens. The pie chart on the right shows the topic distribution.}
\label{fig:query_statistcs}
\end{figure}

Figure \ref{fig:query_statistcs} shows the statistics on our synthetic query data.

\section{Trainning Setups}
\label{sec:training_setups}

Training is performed using the LLaMa Factory \cite{zheng2024llamafactory} framework.

In SFT data without mixup, for a graph $G$, the inputs and outputs are obtained as $(x, y) = (G, p_{\text{opt}})$; with mixup, for $G$, the inputs and outputs include both $(x, y) = (G, p_{\text{opt}})$ and $(G, p_{\text{second}})$. For DPO data, the input $x$, chosen output $y_w$ and rejected output $y_{l}$ are derived as $(G, p_{\text{opt}}, p_{\text{second}})$.

The training loss function of SFT is defined as follows:

\begin{equation}
\setlength{\abovedisplayskip}{5pt}
\setlength{\belowdisplayskip}{5pt}
\mathcal{L}_{\text{SFT}}(\theta) = -\mathbb{E}_{(x,y)\sim \mathcal{D}_{\text{SFT}}} \sum_{t=1}^{|y|} \log P_\theta(y_t | x, y_{<t}).
\end{equation}

The training loss function of DPO is defined as follows:

\begin{equation}
\begin{split}
\mathcal{L}_{\text{DPO}} = {} & -\mathbb{E}_{(x,y_w,y_l) \sim \mathcal{D}_{\text{DPO}}} \\
& \log \sigma\bigg( \beta \bigg( \log \frac{\pi_\theta(y_w|x)}{\pi_{\text{ref}}(y_w|x)} \\
& \quad - \log \frac{\pi_\theta(y_l|x)}{\pi_{\text{ref}}(y_l|x)} \bigg) \bigg).
\end{split}
\end{equation}

The detailed hyperparameter settings are outlined in Table

\begin{table}[!htp]
\setlength{\belowcaptionskip}{-0.2cm}
\setlength{\abovecaptionskip}{0.2cm}
    \centering
    \scalebox{1.}{
    \begin{tabular}{l|c}
        \toprule
        \textbf{Name} & \textbf{Value} \\
        \midrule
        cutoff len & 8,192 \\
        epochs & 10 \\
        batch size per device & 1 \\
        gradient accumulation steps & 4 \\
        learning rate & 1e-6 \\
        lr scheduler type & \texttt{cosine} \\
        warmup ratio & 0.1 \\
        bf16 & \texttt{true} \\
        \bottomrule
    \end{tabular}
    }
    \label{tab:hyperparameters}
    \vspace{-1mm}
    \caption{Detailed training hyperparameters.}
\end{table}


\begin{table*}[htbp]
\setlength{\belowcaptionskip}{-0.2cm}
\setlength{\abovecaptionskip}{0.2cm}
\small
\centering
\resizebox{\textwidth}{!}{
\begin{tabular}{l|cc|cc|cc|cc}
\toprule
\multirow{2}{*}{\textbf{Model}} & \multicolumn{2}{c|}{\textbf{Success Rate}} & \multicolumn{2}{c|}{\textbf{Optimal Rate}} & \multicolumn{2}{c|}{\textbf{Time Ratio}} & \multicolumn{2}{c}{\textbf{Cost Ratio}} \\
& \textit{r} & \textit{m} & \textit{r} & \textit{m} & \textit{r} & \textit{m} & \textit{r} & \textit{m} \\
\midrule
Claude 3.5 Sonnet & -0.96 & -0.99 & -0.82 & -0.82 & 0.92 & 1.02 & 0.94 & 0.99 \\
GPT-4o   & -0.99 & -1.02 & -0.94 & -0.97 & 0.99 & 1.09 & 0.98 & 1.1 \\
Llama-3.1-8B-Instruct       & -0.95 & -0.96 & -0.95 & -1.01 & 0.79 & 0.92 & 0.93 & 0.97 \\
Llama-3.1-8B-Instruct-Trained & -0.94 & -1.04 & -0.89 & -0.86 & 0.94 & 1.04 & 0.93 & 1.0 \\
Qwen2.5-7B-Instruct        & -0.96 & -0.98 & -0.96 & -0.95 & 0.96 & 0.97 & 0.66 & 0.66 \\
Qwen2.5-7B-Instruct-Trained & -0.99 & -0.98 & -0.80 & -0.82 & 0.81 & 0.76 & 0.96 & 0.98 \\
\bottomrule
\end{tabular}
}
\caption{Correlation coefficients (r) and slopes (m) between metrics and node counts.}
\label{tab:model_metrics_node}
\end{table*}

\begin{table*}[t]
\setlength{\belowcaptionskip}{-0.2cm}
\setlength{\abovecaptionskip}{0.2cm}
\small
\centering
\resizebox{\textwidth}{!}{
\begin{tabular}{l|l|cc|cc|cc|cc}
\toprule
\multirow{2}{*}{\textbf{Model}} & \multirow{2}{*}{\textbf{Node Count}} & \multicolumn{2}{c|}{\textbf{Success Rate}} & \multicolumn{2}{c|}{\textbf{Optimal Rate}} & \multicolumn{2}{c|}{\textbf{Time Ratio}} & \multicolumn{2}{c}{\textbf{Cost Ratio}} \\
& & \textit{r} & \textit{m} & \textit{r} & \textit{m} & \textit{r} & \textit{m} & \textit{r} & \textit{m} \\
\midrule
\multirow{2}{*}{Claude 3.5 Sonnet} 
& 10  & 0.27 & 0.15 & -0.44 & -0.28 & 0.50 & 0.26 & 0.43 & 0.23 \\
& 30  & -0.74 & -0.56 & -0.71 & -0.55 & 0.81 & 0.59 & 0.78 & 0.59 \\
\cmidrule{1-10}
\multirow{2}{*}{Llama-3.1-8B-Instruct-Trained} 
& 10  & 0.05 & 0.03 & -0.28 & -0.17 & 0.58 & 0.34 & 0.45 & 0.24 \\
& 30  & -0.39 & -0.29 & -0.62 & -0.37 & 0.45 & 0.34 & 0.60 & 0.43 \\
\bottomrule
\end{tabular}
}
\caption{Correlations (r) and slopes (m) between edge variations and metrics with node counts.}
\label{tab:model_metrics_edge}
\end{table*}

\section{Edge variation Status}

\begin{figure}[th]
\centering
\includegraphics[width=\linewidth]{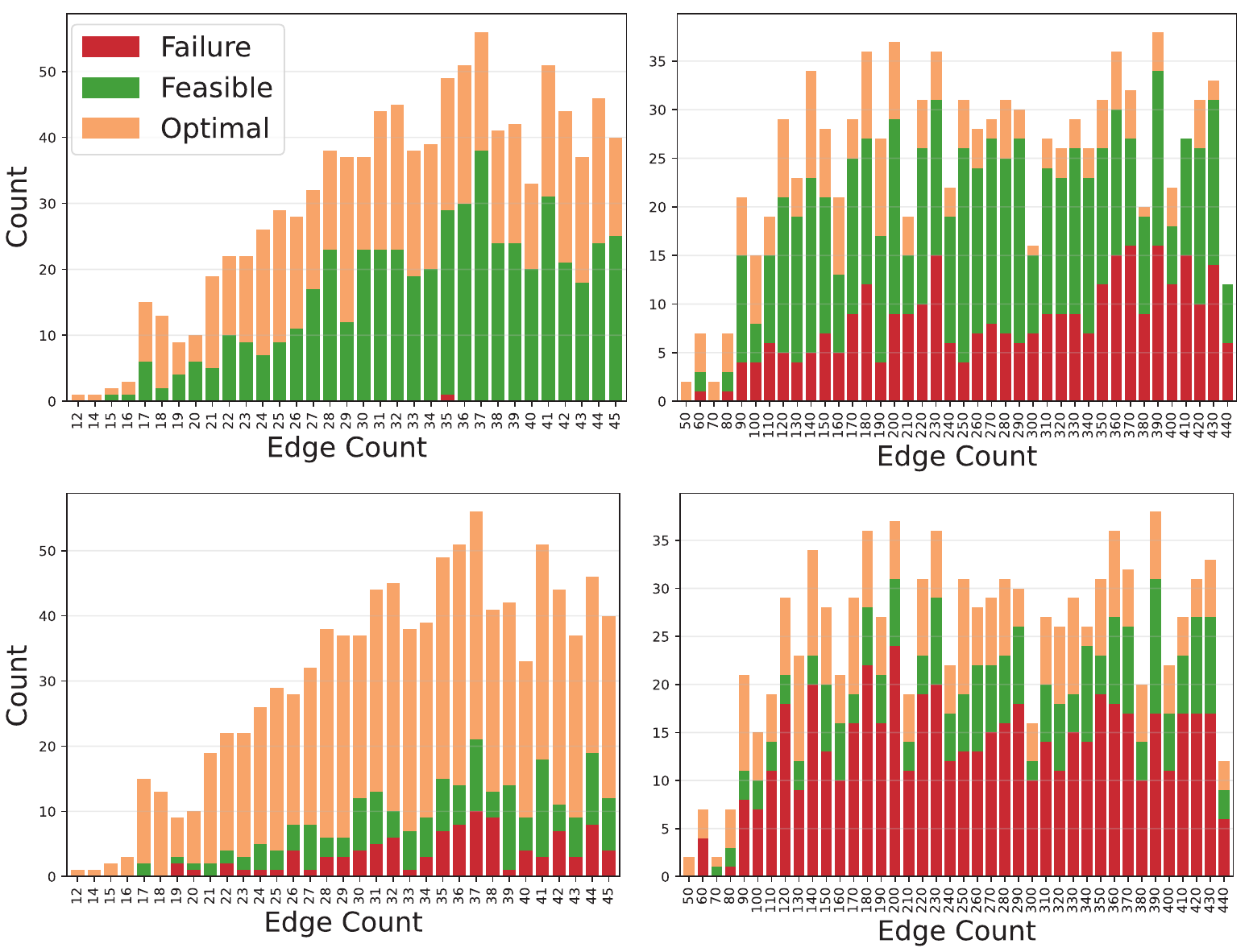}
\caption{Claude and our trained Llama performance across different edge counts. The vertical axis represents corresponding number of cases of each state(fail, feasible, optimal). The horizontal axis represents the number of edges segmented at certain intervals.}
\label{fig:status_results}
\end{figure}

\section{Query Example}
\label{query_example}

A query and corresponding plan in a real-life scenario is shown in \ref{sec:examples}.

\section{Supplementary of the Experiments}
\label{sec:supplementary_of_the_experiments}

Tabel \ref{tab:model_metrics_node} and \ref{tab:model_metrics_edge} show the absolute value of correlation coefficients and slopes of normalized four metrics with changes in the number of nodes and edges.

Metrics are scaled to [0,1] range using min-max normalization:
\begin{equation}
y_{\text{norm}} = \frac{y - y_{\min}}{y_{\max} - y_{\min}}
\end{equation}
where $y$ represents raw values of: node and edge counts, success counts, optimal counts, average time ratios, and average cost ratios.

Pearson correlation coefficients ($r$) between node and edge counts ($X$) and metrics ($Y$) are calculated as:
\begin{equation}
r_{XY} = \frac{\sum_{i=1}^n (x_i - \bar{x})(y_i - \bar{y})}{\sqrt{\sum_{i=1}^n (x_i - \bar{x})^2} \sqrt{\sum_{i=1}^n (y_i - \bar{y})^2}}.
\end{equation}

The slope ($m$) of the best-fit line is computed using linear regression:
\begin{equation}
m = \frac{n\sum x_iy_i - \sum x_i \sum y_i}{n\sum x_i^2 - (\sum x_i)^2}.
\end{equation}

All results are rounded to 2 decimal places for final reporting.

For a given plan $P$ composed of sub plans $p$, the time ratio of parallel execution to sequential execution can be calculated as: 

\begin{equation}
    \text{Ratio} = \frac{\Omega_{\text{time}}(P)}{\sum\limits_{p\in P} \tau_{p}}.
\end{equation}

\end{document}